%% file: main.tex
\documentclass{article}
\usepackage{iclr2027_conference,times}

\usepackage[utf8]{inputenc}
\usepackage[T1]{fontenc}
\usepackage{hyperref}
\usepackage{url}
\usepackage{booktabs}
\usepackage{amsmath}
\usepackage{amssymb}
\usepackage{graphicx}
\usepackage{subcaption}
\usepackage{xcolor}
\usepackage{xspace}
\usepackage{pifont}
\usepackage{microtype}

\newcommand{\method}{\textbf{AnswerMap}\xspace}
\newcommand{\cmark}{\ding{51}}
\newcommand{\xmark}{\ding{55}}

\title{\method: Faithful Spatial Interpretability of VLMs from Answer Posteriors}

\iclrfinalcopy 
\author{%
\makebox[\textwidth][c]{Mohamed Eltahir$^{1*}$\hspace{2.0em}Fardows Adam$^{1*}$\hspace{2.0em}Duaa M. Tahir$^{1*}$\hspace{2.0em}Lama Alamoudi$^{1}$\hspace{2.0em}Sana Ammar$^{1}$}\\[0.25em]
\makebox[\textwidth][c]{\textbf{Atheer A. Alboloshi}$^{1}$\hspace{2.0em}\textbf{Jory Albluey}$^{1}$\hspace{2.0em}\textbf{Tanveer Hussain}$^{2\ddagger}$\hspace{2.0em}\textbf{Naeemullah Khan}$^{1\S}$}\\[1.0em]
\makebox[\textwidth][c]{$^{1}$King Abdullah University of Science and Technology (KAUST), Thuwal, Saudi Arabia}\\
\makebox[\textwidth][c]{$^{2}$Department of Computer Science, Edge Hill University, Ormskirk, England}\\[0.3em]
\makebox[\textwidth][c]{\small\texttt{\{mohamed.hamid, fardoos.ahmad, duaa.tahir, lama.alamoudi, sana.taffour,}}\\
\makebox[\textwidth][c]{\small\texttt{atheer.alboloshi, jory.albluey, naeemullah.khan\}@kaust.edu.sa}}\\
\makebox[\textwidth][c]{\small\texttt{hussaint@edgehill.ac.uk}}
}

\begin{document}

\vspace*{-30pt}
\maketitle
\lhead{Preprint.}

{\renewcommand{\thefootnote}{\fnsymbol{footnote}}
\footnotetext[1]{Equal contribution.}
\footnotetext[3]{Corresponding author.}
\footnotetext[4]{Principal Investigator (PI).}
\footnotetext{Code: https://github.com/sanataff/AnswerMap}
}

\vspace{-20pt}

\begin{figure}[h]
\centering
\includegraphics[width=0.9\linewidth]{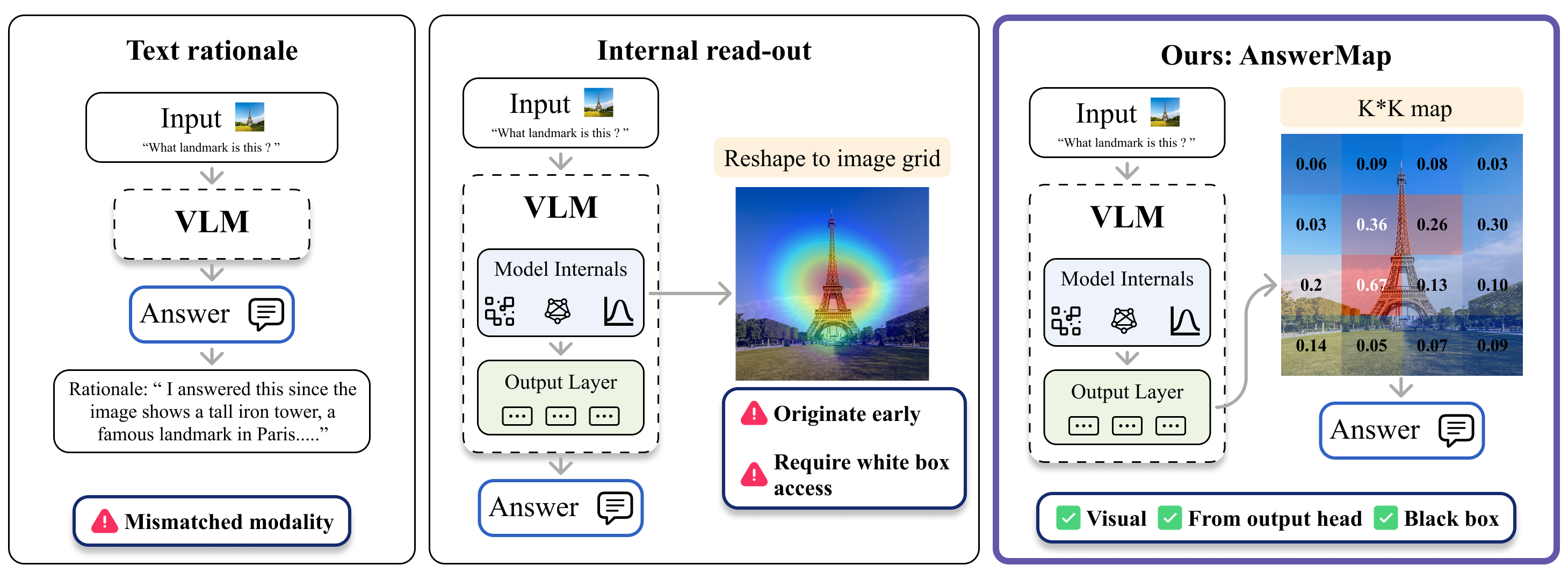}
\caption{Three ways to explain one frozen VLM's answer. A text rationale is in a mismatched modality. An internal read-out is in the right modality but originates too early and needs the weights. \method{} reads the output head one band at a time, the map is visual and black-box, and a fixed read-out computes the answer from it.}
\label{fig:teaser}
\end{figure}

\vspace*{-0.4cm}
\begin{abstract}
\input{sections/abstract}
\end{abstract}


\begin{figure}[t]
\centering
\includegraphics[width=\linewidth]{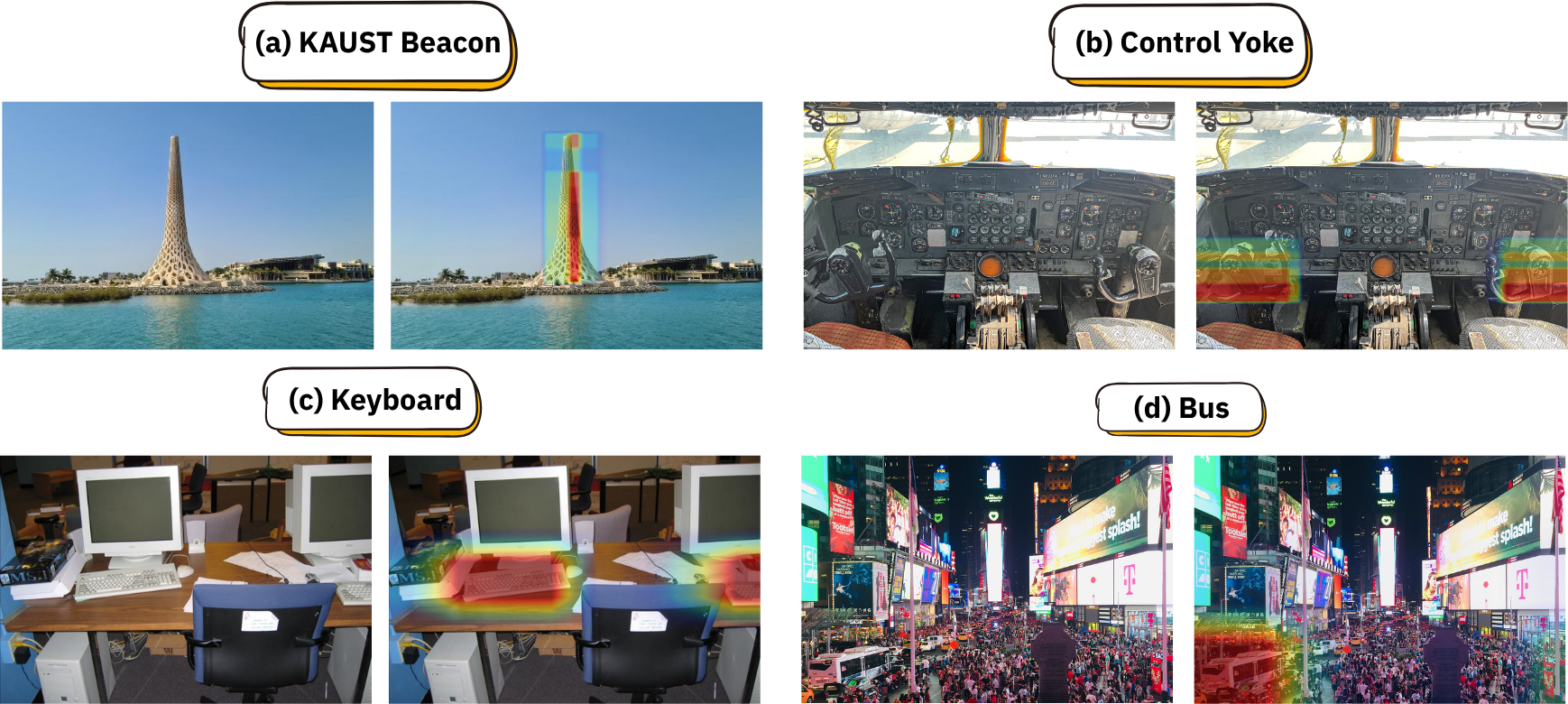}
\caption{\method{} (multigrid) from GPT-6-sol through its API, one
query each.}
\label{fig:api_examples}
\end{figure}

\input{sections/introduction}

\begin{table}[t]
\centering
\small
\caption{Positioning. Any answer: the rationale exists for non-word answers
such as coordinates. Cell score:
what a cell's value measures (where attention routed, how strongly a word
activates, whether the cell is necessary when removed from the full image,
or whether a band shown alone suffices). Black box: needs only the model's
output token probabilities. Passes: forward
passes for one $K\times K$ map.}
\label{tab:positioning}
\setlength{\tabcolsep}{4pt}
\resizebox{\textwidth}{!}{%
\begin{tabular}{lccccc}
\toprule
\textbf{Rationale} & Spatial & Any answer & Cell score & Black box & Passes per map \\
\midrule
Words (text generation)          & \xmark & \cmark & none        & \cmark & 1 \\
Attention                        & \cmark & \cmark & routing     & \xmark & 1 \\
Gradient relevancy (e.g., T-MM)  & \cmark & \cmark & routing     & \xmark & 1 \\
Token activation (e.g., TAM)     & \cmark & \xmark & activation  & \xmark & 1 \\
Occlusion                        & \cmark & \cmark & necessity   & \cmark & $K^2$ \\
\midrule
\method                          & \cmark & \cmark & sufficiency & \cmark & $2K$ \\
\bottomrule
\end{tabular}}
\end{table}

\input{sections/related_work}

\input{sections/methodology}

\input{sections/experiments}
\input{sections/applications}
\input{sections/ablations}

\input{sections/conclusion}



\subsubsection*{Acknowledgments}
We are grateful to the KAUST Academy for its generous
support. For computer time, this research used Ibex managed by the
Supercomputing Core Laboratory at King Abdullah University of Science \&
Technology (KAUST) in Thuwal, Saudi Arabia.

\bibliography{bibliography}
\bibliographystyle{iclr2027_conference}

\newpage
\appendix
\input{sections/appendix}

\end{document}

%% file: sections/abstract.tex
When a VLM answers a visual query, current interpretability tools rely on text rationales, which use a mismatched modality, or on internal read-outs, which originate too early to reflect the final output and require white-box access to the model. We introduce
\method{}, a training-free, task-agnostic, black-box visual rationale constructed from
the output head. The image is cut into $K$ row and $K$ column bands, each
shown alone to the frozen model along with the query in the format of a
yes/no relevance question. The outer product of the row and column ``yes''
posteriors gives the query-conditioned spatial map. Crucially, by defining
a fixed read-out $R$ (e.g., expectation, maximum) on top of \method{}, we
can derive continuous outputs like location natively. This bypasses the
reliance on discrete text tokens for continuous-output tasks and guarantees
an image-dependent answer by construction. However, a rationale can be
confabulated, so we validate \method{} across four models and three query
distributions with two tests: (a) agreement with the model's own generated
point and (b) deletion of the map's region. The map lands where the model
points (AUC 0.85 against 0.38 for attention), and deleting its region flips
53\% of correct answers (against 19\% for
attention's). Beyond establishing faithfulness, we demonstrate the
map's task-agnostic utility through three distinct read-outs: its maximum
flags hallucinated objects without generation, its expectation localizes
correctly when the model's own pointing fails, and its top-mass region, fed
back as a crop, fixes half of the model's wrong answers. \method{} thus offers a new lens on VLM interpretability and, through its read-outs, a new output interface for visual tasks beyond text tokens. 

%% file: sections/introduction.tex
\section{Introduction}\label{sec:intro}

When a vision language model (VLM) answers a visual query, the natural
check is spatial: which part of the image did the answer come from. A
radiologist reading a generated report or an auditor checking a claim
needs that check. Providing it is the job of interpretability tools.

Current interpretability tools for VLMs suffer from distinct structural limits (Figure~\ref{fig:teaser}). Text rationales come from the same head as the answer but operate in a mismatched modality. They name no explicit region, cannot be tested against the image, and often fail to reflect what actually drove the prediction
\citep{turpin2023,edct2510}. Internal read-outs, attention and gradient
attribution, are in the right modality but originate earlier in the network than where the answer is generated. The relationship between their values and the final output is notoriously unclear, a verdict measured first on text models and now on VLMs \citep{jain2019attention,attnfaith2609}, and much of their mass sits on sink tokens the answer does not need \citep{sink2503}. A third line of work, perturbation \citep{zeiler2014,rise2018}, avoids these limits. It is visual, reads the final output, and requires no internal access. However, standard occlusion scores one cell at a time by removing it from the full image. Consequently, each score is merely the difference between two near-certain answers, a cell scores nothing if the rest of the image covers for it, and the method scales quadratically, paying one forward pass per cell.

We build on this third line but change what each cell's score sees. \method{}, a training-free, task-agnostic visual rationale (Figure~\ref{fig:method}), does not read the model's words or its internals. Instead, it asks the model about each part of the image and records the answer probability. The image is cut into $K$ row and $K$ column bands, and each band is shown to the frozen model in isolation alongside one yes/no question about the query. Thus, every score is a full-contrast recognition judgment of a region on its own, rather than a measure of its removal from context. The outer product of the row and
column "yes" posteriors is a $K\times K$ query-conditioned map, $O(K)$
queries rather than $O(K^2)$.

Crucially, the rationale does more than explain. Applying a fixed read-out $R$ to the map derives the answer directly, such as taking its expectation for a location or its maximum for grounding strength. This has two major structural consequences. First, the output becomes continuous: a coordinate rather than a discrete token. Second, because every entry is a posterior derived from a single strip shown in isolation, there is no unconstrained generation step where the language prior can take over. So, reliance on the image is guaranteed by construction.

Because any rationale can be confabulated, we introduce two ground-truth-free tests for spatial explanations: (a) the model's own generated point should land where the map is high, and (b) deleting the map's peak region should break the answer. \method{} passes both tests across four models. In contrast, attention fails both, and the strongest white-box baseline trails on each. Finally, three different read-outs applied to the same map actively solve downstream tasks: flagging hallucinated objects, localizing when the model's own pointing fails, and fixing the model's prior mistakes through a targeted crop.

\textbf{Contributions.} (1) \method{}, a black-box spatial rationale from
answer posteriors at $2K$ queries, with a multigrid product to capture different types of context. (2) Read-outs on the map as an output interface.
(3) A two-test, ground-truth-free faithfulness protocol for any spatial
explanation of any VLM.

%% file: sections/related_work.tex
\section{Related Work}\label{sec:related}

\textbf{Text rationales.}
The default explanation of a VLM answer is the model's own words. However, a model's stated reasoning can misrepresent the
cause of its prediction \citep{turpin2023}, text supplies most of a VL
decoder's output \citep{parcalabescu2404}, a model that writes ``let me
check the figure again'' rarely notices a swapped figure
\citep{seeingsaying2605}, and VLM explanations only partly survive
counterfactual tests of the concepts they cite \citep{edct2510}. The fundamental limit across all four failure modes is the modality: a sentence names no explicit region, so nothing in it can be tested against the image.

\textbf{Attention as explanation.}
Attention weights, raw or propagated by rollout \citep{rollout2020}, are the
most common lens on where a transformer looks. Text-domain intervention studies were the first to question this lens, showing that very different attention distributions can yield the same prediction \citep{jain2019attention}, and that attention predicts input importance only noisily \citep{serrano2019attention}. For
VLMs the signal is further distorted. Image tokens receive a vanishing
share of attention in deep layers \citep{fastv2403}, high attention lands
on sink tokens irrelevant to the text \citep{sink2503}, and attention systematically under-allocates to the regions a question actually needs \citep{lostattn2605}. Furthermore, grounding from attention requires first finding the few specific heads that localize \citep{attnheads2503}, and its faithfulness varies by input, often exhibiting a mode where no attended region is individually necessary \citep{attnfaith2609}.
The limit here is structural: an attention weight records how much a position was read, but nothing directly ties that number to the final answer. A cell score that is
itself an answer about the cell would not inherit this problem.

\textbf{Gradient, relevancy, and activation attribution.}
Grad-CAM \citep{gradcam2017}, integrated gradients \citep{ig2017}, and Transformer-MM relevancy propagation (T-MM) \citep{chefer2021} route the output's gradient back to the input. Recent methods also condition directly on the generated answer: GLIMPSE fuses gradient-weighted attention across layers \citep{glimpse2506}, and TAM projects image-token states onto a generated word through the output layer \citep{tam2506}. On vision transformer
classifiers a salience-based faithfulness test finds every method in the
family only moderately faithful, Grad-CAM and integrated gradients low,
plain LRP near the random floor, and gradient-weighted attention
aggregated across layers, T-MM, at the top \citep{wu2024faithfulness}. Additionally, multi-encoder VLM architectures make gradient methods notoriously difficult to apply \citep{wherelook2503}.
Crucially, all of these methods require white-box access to the model weights, a ceiling the mechanistic literature explicitly acknowledges as its own limitation \citep{mann2026wherereliability}.

\textbf{Black-box perturbation.}
Occlusion \citep{zeiler2014} and RISE \citep{rise2018} estimate importance
by masking the input and watching the output, which needs no internals and
can be tested by the same intervention. The closest method to our setting
optimizes a deletion-validated heatmap for VLM answers
\citep{wherelook2503}, but it needs gradients and the generated answer. For purely black-box approaches, computational cost is the bottleneck: they require one forward pass per occluder position (e.g., 4,000 to 8,000 masked passes per map for RISE). Furthermore, each score represents the marginal effect of one cell against a full image that still displays context everywhere else. A signal that judges a region on its own, combined with a factorization that keeps the pass count linear in the grid, would preserve the black-box access while eliminating the prohibitive cost.

\textbf{Low-rank factorization and answer-space probing.}
Writing a matrix as an outer product of two vectors is the cheapest
structure a matrix can have, and it is a workhorse wherever a full matrix
is too expensive to learn or measure, from separable filters to low-rank
adaptation of weight updates \citep{hu2021lora}. GridProbe brought the same
structure to inference: it arranges video frames on a $K\times K$ grid,
asks a frozen VLM $2K$ questions about the rows and the columns, and reads
the outer product of the answer confidences as a frame importance map,
$O(K)$ passes for $K^2$ candidates \citep{gridprobe2605}. An image is a
grid too. If the same $2K$ questions read a map on an image, the map is no
longer a means to select inputs but the explanation itself, at the cost
perturbation cannot reach.

Across these threads, text rationales cannot be tested against the image, internal read-outs are unreliable even when weights are available, and perturbation is testable and black-box but pays a steep pass-per-cell cost for a signal diluted by surrounding context. What is missing is a spatial rationale whose cell scores are direct answers, read entirely from the output head at a linear cost.
The following section builds it, and Table~\ref{tab:positioning}
summarizes the positioning.

%% file: sections/methodology.tex
\section{\method{}}\label{sec:method}

\begin{figure}[t]
\centering
\includegraphics[width=\linewidth]{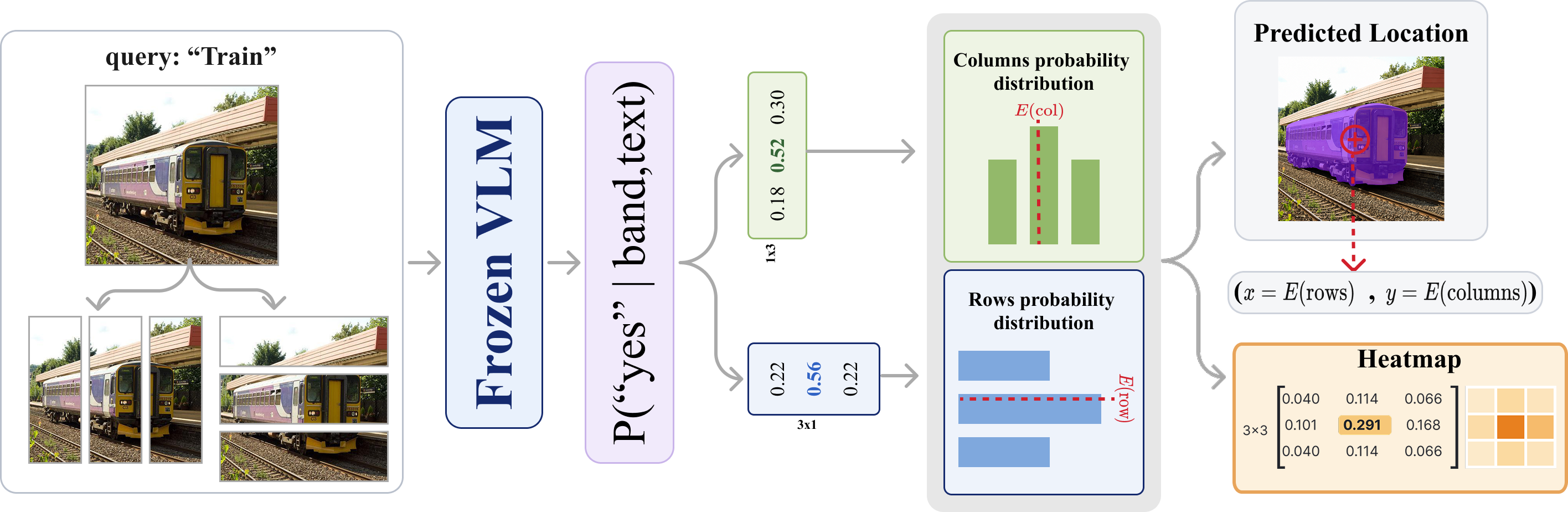}
\caption{The \method{} operator. $K$ row and $K$ column band questions
fill a $K\times K$ map through the outer product, and a read-out computes
the answer from the map.}
\label{fig:method}
\end{figure}

\subsection{What the probe reads}\label{sec:method:op}

Let $x$ be an image and $q$ a free-text query. We cut $x$ into $K$
horizontal bands and $K$ vertical bands. Each band is cropped out and shown
alone to the frozen model, as one strip image with no index and no
surrounding image, with one binary question, ``is \{$q$\} in this band''.
The answer is never read from generated text. Let $z_{\mathrm{yes}}$ and
$z_{\mathrm{no}}$ be the model's first answer-token logits for the two
labels. The yes
posterior of row band $i$ is
\begin{equation}
p_i^{\mathrm{row}} = \frac{\exp z_{\mathrm{yes}}}
{\exp z_{\mathrm{yes}} + \exp z_{\mathrm{no}}}\,,
\qquad i = 1,\dots,K,
\end{equation}
and likewise $p_j^{\mathrm{col}}$ for column band $j$. The two marginals
combine into a $K\times K$ map by the outer product
\begin{equation}
M_{ij} = p_i^{\mathrm{row}}\, p_j^{\mathrm{col}} ,
\label{eq:outer}
\end{equation}
where cell $(i,j)$ is the intersection of row band $i$ and column band $j$.
Intuitively, a cell is important only if both the row and the column
through it draw a confident yes. One confident marginal gives partial
weight, neither gives none. An $8\times8$ map costs $2K{=}16$ forward
passes, against 65 for occlusion at the same grid, and
the only access it needs is the first-token logprobs of two labels. The
bands are cut in pixel space, so the operator never touches the model's
token layout: dynamic resolution, tiling, and token merging, which an
internal read-out must invert to reach pixels, are invisible to it. It
does need a model that accepts non-square inputs, natively or by tiling.
The signal also differs from occlusion's, not only its cost. Occlusion asks whether a cell is necessary given the rest of the
image, a signal that redundancy hides. The probe asks whether a band alone
suffices, so each answer is one recognition judgment, and the
factorization turns $2K$ of them into a $K\times K$ map. 

\textbf{The read-out.} The map is the rationale. The answer is a
\textbf{read-out} $R(M)$, a fixed function of the map, computed rather
than generated. We use two in the tests and a third in
Section~\ref{sec:applications}. Let $c_{ij}$ be the centre of cell $(i,j)$
in image coordinates, and let $m = \min_{i,j} M_{ij}$ be the smallest
entry of the map. The \textbf{expectation} is the weighted mean of the
cell centres,
\begin{equation}
\hat\ell(M) = \sum_{i,j} w_{ij}\, c_{ij},
\qquad
w_{ij} = \frac{M_{ij} - m}{\sum_{i',j'} (M_{i'j'} - m)},
\label{eq:expectation}
\end{equation}
a continuous sub-cell location. Subtracting $m$ removes the map's floor,
which puts the probe on the same zero-floor footing every baseline map has
by construction. The \textbf{maximum} $\max_{ij} M_{ij}$ is a scalar
grounding strength, high when the model commits to a region, read on the
raw map of Equation~\ref{eq:outer}, whose entries are products of
posteriors.
For a spatial query the read-out is the answer. For a question answered in
words the map is the rationale beside the answer, and
Section~\ref{sec:results} tests both.

\textbf{Why the image must matter.} A VLM can answer a visual question
from its language prior alone, and a fluent answer does not show when it
did \citep{seeingsaying2605,parcalabescu2404}. Under generation the model
sees the image and the question together, and nothing in the emitted text
records what it drew on. The guarantee against this is a property of the
read-out. Under
$R$ the strips differ only in content: no index, no surrounding image, the
same question. Whatever the language prior contributes, it contributes to
every strip alike, since nothing lets it tell one strip from another, so
it can raise or lower the whole map but cannot move mass across it. Any location or peak a
read-out computes therefore comes from the strips, and reliance on the
image is guaranteed by construction.

\begin{figure}[t]
\centering
\includegraphics[width=\linewidth]{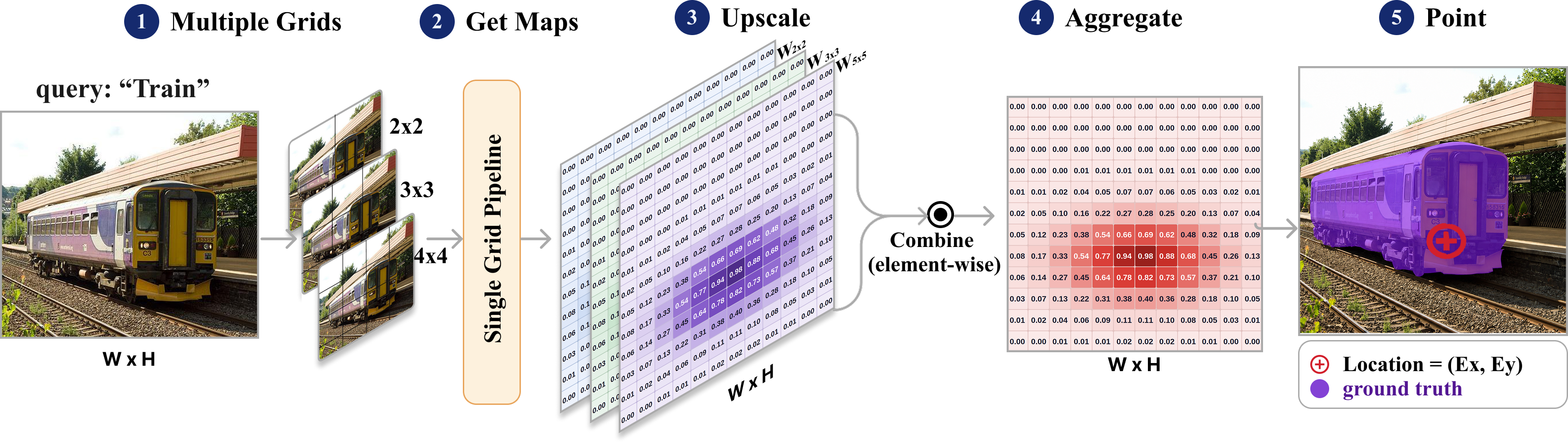}
\caption{Multigrid product. Coprime grids are probed separately, upsampled
to a shared grid, and combined, so a region survives only if every band
width endorses it.}
\label{fig:multigrid}
\end{figure}

\begin{figure}[t]
\centering
\includegraphics[width=0.6\linewidth]{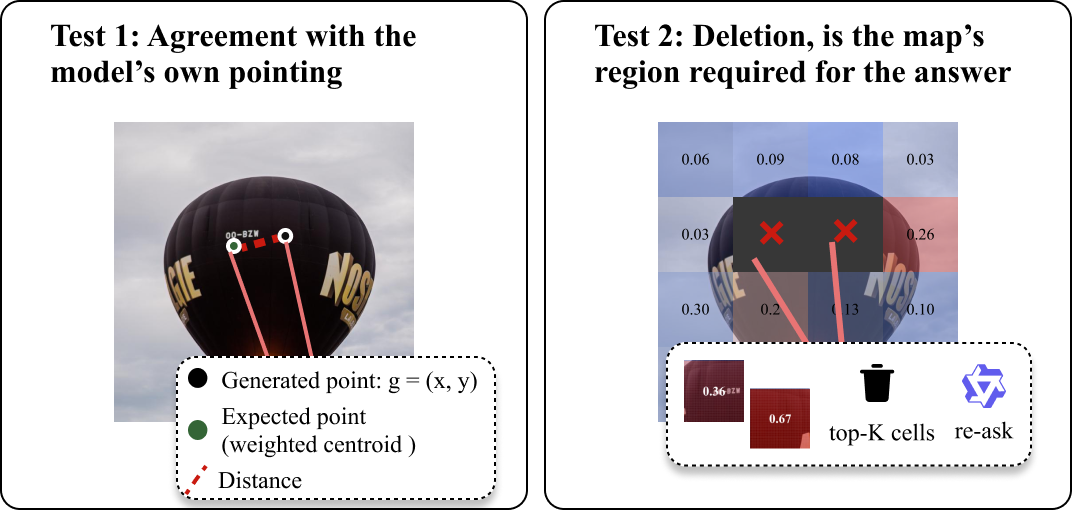}
\caption{The two-test faithfulness protocol. Left, agreement, score the full
map at the model's own generated point. Right, deletion, remove the map's
top-mass region against a matched random region and re-ask.}
\label{fig:tests}
\end{figure}

\subsection{Multigrid product}\label{sec:method:mg}

A single grid fixes one band width, and the band width fixes what each
question can see. A coarse grid shows the model wide bands, so each answer
is judged with context and the map captures global structure. A fine grid shows thin bands, so each answer is local and
precise. We therefore probe several coarse and fine grids,
upsample each map by nearest neighbour to a shared $64\times64$ grid,
min-max normalize each map to $[0,1]$, and multiply them elementwise
(Figure~\ref{fig:multigrid}). The product is a product of experts, a soft
intersection under which a location survives only if every view endorses
it, so a region must be supported both in context and in detail. The
product also gains resolution: a row profile from $K_1$ bands times one
from $K_2$ bands is piecewise constant on the union of their boundaries,
so coprime sizes, which share no boundary, give $K_1{+}K_2{-}1$ pieces
from $K_1{+}K_2$ questions, each still wide enough to answer, whereas
nested sizes such as $\{2,4,8\}$ repeat boundaries and add none. The map
stays separable, a row profile times a column profile, so the product
sharpens one region and does not add a second.

\subsection{The two tests}\label{sec:method:tests}

A map elicited from the model's own answers could be confabulation, so we measure its faithfulness with two
ground-truth-free tests that apply to any spatial explanation
(Figure~\ref{fig:tests}).

\textbf{Test 1, agreement.} The map claims to read the model's belief
about where the query is. The model can also state that belief itself, by
pointing at the query through its native grounding prompt. If the map is
faithful, the generated point is a sample from the belief the map
describes and should fall where the map is high. The test therefore
scores the map at the model's own point, against the model rather than
against ground truth.

\textbf{Test 2, deletion.} The map claims that the answer depended on the
region it names. If the map is faithful, removing that region from the
image should break a correct answer, and removing a random region of the
same size should not. The test therefore deletes the
map's region, re-asks the question, and compares the answer change rate
with the random control.

%% file: sections/experiments.tex
\section{Experiments}\label{sec:results}

\subsection{Setup}\label{sec:results:setup}

All experiments use frozen open-weight models in bfloat16 with greedy
decoding and no training: Qwen3-VL-4B-Instruct \citep{qwen3vl} as the
primary model, replicated on Qwen3-VL-8B, Qwen3-VL-30B-A3B, and
InternVL3.5-8B \citep{internvl35}. The single $K{=}8$ grid is the default
probe, and the coprime pair $\{3,5\}$, which costs the same 16 queries, is
reported beside it. Probes use a 512-pixel long side, generation and deletion use 1024 (probing at 1024 instead leaves the agreement
scores unchanged, Appendix~\ref{app:config}). Baselines are zero-shot,
calibration-free,
and answer-free. Attention enters three ways: at its best layer, at the
last layer, and by rollout \citep{rollout2020}. The best layer is chosen
once per model by sweeping all decoder layers on 200 RefCOCOg examples,
then fixed for every benchmark and both tests. The 200 are among the 800
RefCOCOg examples evaluated, an overlap that can only favor attention. T-MM relevancy
\citep{chefer2021} is the white-box representative. Occlusion
\citep{zeiler2014} runs on the same $8\times8$ grid, one grey cell at a
time. A random map is the floor. Within each table all methods are scored
on the same examples. TAM \citep{tam2506} is white-box and explains a generated token, so it
runs in its own regime: in Test 1 the model names the queried object and
TAM explains the name, with the best sub-token counted as in its Obj-IoU,
in Test 2 it explains the generated answer. Prompts are in Appendix~\ref{app:prompts} and the full configuration in
Table~\ref{tab:config}.

\textbf{Metrics.} Test 1 scores the full map at the model's generated
point $g$ with the saliency-canon metrics \citep{bylinskii2018different}: NSS, the
z-scored map value at $g$ (chance 0), and AUC, the probability that the
map ranks the cell containing $g$ above a random cell (chance 0.5). Both
score the map directly, so multimodal maps are treated fairly, and each
map is scored at its native resolution. Test 2 pools every map to the shared $8\times8$ grid, deletes its 8 highest-valued cells
(12.5\% of the image, not necessarily contiguous) by setting them to grey,
and reports the answer change rate on questions the model answers
correctly with the full image. Controls delete a random region of the
same cell count, drawn uniformly or disjoint from method regions.

\subsection{Test 1: the map lands where the model points}\label{sec:results:agreement}

Table~\ref{tab:agreement} reports Test 1 on referring expressions
RefCOCOg \citep{refcocog2016}, appearance-only expressions RefCOCO+ \citep{refcocoplus2016}, and anomaly descriptions CAVE \citep{cave2510}, with the same ordering on
all three. Qualitative maps are in Appendix~\ref{app:qualitative}.

\begin{table}[t]
\centering
\small
\caption{Test 1. Does the map land where the model points? NSS and AUC at
the model's own generated point (chance 0 and 0.5). RefCOCOg n=800, RefCOCO+
n=800, CAVE n=334. TAM is answer-conditioned and runs in its own regime
(Section~\ref{sec:results:setup}). Controls: the probe run on a blank
image, and on the image of another example, with the model's generated
point held fixed.}
\label{tab:agreement}
\begin{tabular}{lcccccc}
\toprule
 & \multicolumn{2}{c}{RefCOCOg} & \multicolumn{2}{c}{RefCOCO+} & \multicolumn{2}{c}{CAVE} \\
\textbf{Method} & NSS & AUC & NSS & AUC & NSS & AUC \\
\midrule
\method{} ($K{=}8$)        & 1.36 & 0.824 & 1.44 & 0.817 & \textbf{1.73} & 0.828 \\
\method{} ($\{3,5\}$, equal cost) & \textbf{1.56} & \textbf{0.853} & \textbf{1.52} & \textbf{0.840} & 1.70 & \textbf{0.845} \\
\midrule
Attention (best layer)     & $-0.22$ & 0.380 & $-0.18$ & 0.432 & $-0.21$ & 0.409 \\
Attention (last layer)     & $-0.15$ & 0.464 & $-0.11$ & 0.490 & $-0.17$ & 0.456 \\
Attention rollout          & $-0.16$ & 0.457 & $-0.15$ & 0.475 & $-0.13$ & 0.543 \\
Relevancy T-MM             & 0.33 & 0.656 & 0.62 & 0.689 & 0.44 & 0.668 \\
Token activation TAM (own regime) & 0.77 & 0.693 & 0.74 & 0.694 & 0.60 & 0.626 \\
Occlusion (65 queries)     & 1.04 & 0.649 & 1.14 & 0.654 & 1.12 & 0.644 \\
\midrule
Random map                & $-0.05$ & 0.486 & $-0.02$ & 0.493 & $-0.01$ & 0.497 \\
Probe map, blank image    & 0.00 & 0.500 & 0.00 & 0.500 & 0.00 & 0.500 \\
Probe map, swapped image  & $-0.02$ & 0.475 & $-0.02$ & 0.490 & $-0.03$ & 0.494 \\
\bottomrule
\end{tabular}
\end{table}

\begin{table}[t]
\centering
\small
\caption{Test 2. Does deleting the map's region break the answer? Answer
change rate (\%) on questions answered correctly with the full image.
TextVQA n=400, GQA n=800. Right columns: blur instead of grey, and
non-binary questions only, those whose answer is not yes or no. TAM is
answer-conditioned and runs in its own regime.}
\label{tab:deletion}
\begin{tabular}{lcccc}
\toprule
\textbf{Method} & TextVQA & TextVQA (blur) & GQA & GQA (non-binary) \\
\midrule
\method{} (ours)                  & \textbf{53.4} & \textbf{54.0} & \textbf{22.4} & \textbf{26.2} \\
\midrule
Relevancy T-MM                    & 36.8 & 36.5 & 17.1 & 15.3 \\
Attention (best layer)            & 18.8 & 19.3 & 3.0 & 2.2 \\
Attention (last layer)            & 18.3 & 19.6 & 4.2 & 4.9 \\
Attention rollout                 & 9.5 & 9.5 & 3.4 & 5.3 \\
Token activation TAM (own regime) & 33.2 & 32.7 & 8.8 & 13.9 \\
Occlusion (65 queries)            & 48.8 & 49.0 & 18.2 & 19.2 \\
\midrule
Random region, uniform            & 8.2 & 8.7 & 4.5 & 5.6 \\
Random region, disjoint from all maps & 3.3 & 3.3 & 2.7 & 2.2 \\
\bottomrule
\end{tabular}
\end{table}

The probe leads and the equal-cost multigrid $\{3,5\}$ leads further.
Occlusion, the only other black-box method above chance, has the same
access as the probe and four times its queries, yet trails on both
metrics, so the margin comes from what each cell's score measures, not from black-box access. Attention is at or below chance even at the
best of its 36 layers (Figure~\ref{fig:layers}), for a mechanical reason:
its mass sits on sink tokens irrelevant to the text \citep{sink2503}, so
its expectation barely moves with the query (Figure~\ref{fig:scatter}).
T-MM, the strongest white-box baseline, and TAM land
between the two.

The two control rows test the guarantee of Section~\ref{sec:method:op}
with the generated point held fixed and only the probe input changed. A
blank image gives a flat map and both metrics land exactly at chance. An
image swapped in from another example gives a map that follows that
image: no correlation with this image's point, and an expectation farther
from it than the image centre (Table~\ref{tab:agreement_detail}), so the map is
not neutral but committed to what it was shown. The query alone carries
no location, and the image alone decides where the map goes.

\subsection{Test 2: deleting the map's region breaks the answer}\label{sec:results:deletion}

Table~\ref{tab:deletion} reports Test 2 on TextVQA \citep{textvqa2019},
where answers are text at a location, and GQA \citep{gqa2019}, where
answers are objects and relations. The probe's region breaks the answer at
six times the rate of a uniform random region on TextVQA and five times
on GQA, the gap holds under blur and on the non-binary questions, and the
strongest baselines are occlusion, the method closest in construction to
a deletion test, which trails the probe at four times the queries, and
T-MM among the white-box read-outs, well above attention and well below
the probe.

\subsection{The result holds across scale and family}\label{sec:results:scale}

The probe is stable everywhere with no per-model tuning, while attention's best layer moves with the model. On InternVL occlusion matches the probe's NSS and trails its AUC, at four times the queries.

\begin{table}[!h]
\centering
\small
\caption{Both tests at 8B, 30B, and on a second family. Test 1: the probe
and the strongest baseline run on that model, attention at the model's own
swept best layer. Test 2 on TextVQA: the probe, occlusion, and a uniform
random region of the same size, all measured on the same model and
examples.}
\label{tab:scale}
\begin{tabular}{lcclcc|ccc}
\toprule
& \multicolumn{5}{c|}{Test 1, agreement} & \multicolumn{3}{c}{Test 2, deletion} \\
 & \multicolumn{2}{c}{Probe} & \multicolumn{3}{c|}{Best baseline} & & & \\
\textbf{Model} & NSS & AUC & & NSS & AUC & Probe & Occlusion & Random \\
\midrule
Qwen3-VL-4B      & \textbf{1.36} & \textbf{0.824} & occlusion & 1.04 & 0.649 & \textbf{53.4} & 48.8 & 8.2 \\
Qwen3-VL-8B      & \textbf{1.49} & \textbf{0.823} & attention & 0.33 & 0.634 & \textbf{51.1} & 50.3 & 9.4 \\
Qwen3-VL-30B-A3B & \textbf{1.39} & \textbf{0.813} & occlusion & 1.03 & 0.671 & \textbf{45.8} & 45.0 & 7.4 \\
InternVL3.5-8B   & 1.03 & \textbf{0.768} & occlusion & 1.03 & 0.699 & \textbf{50.4} & 50.0 & 9.6 \\
\bottomrule
\end{tabular}
\end{table}

\subsection{The operator runs on a closed model}
\begin{figure}[!h]
\centering
\includegraphics[width=\linewidth]{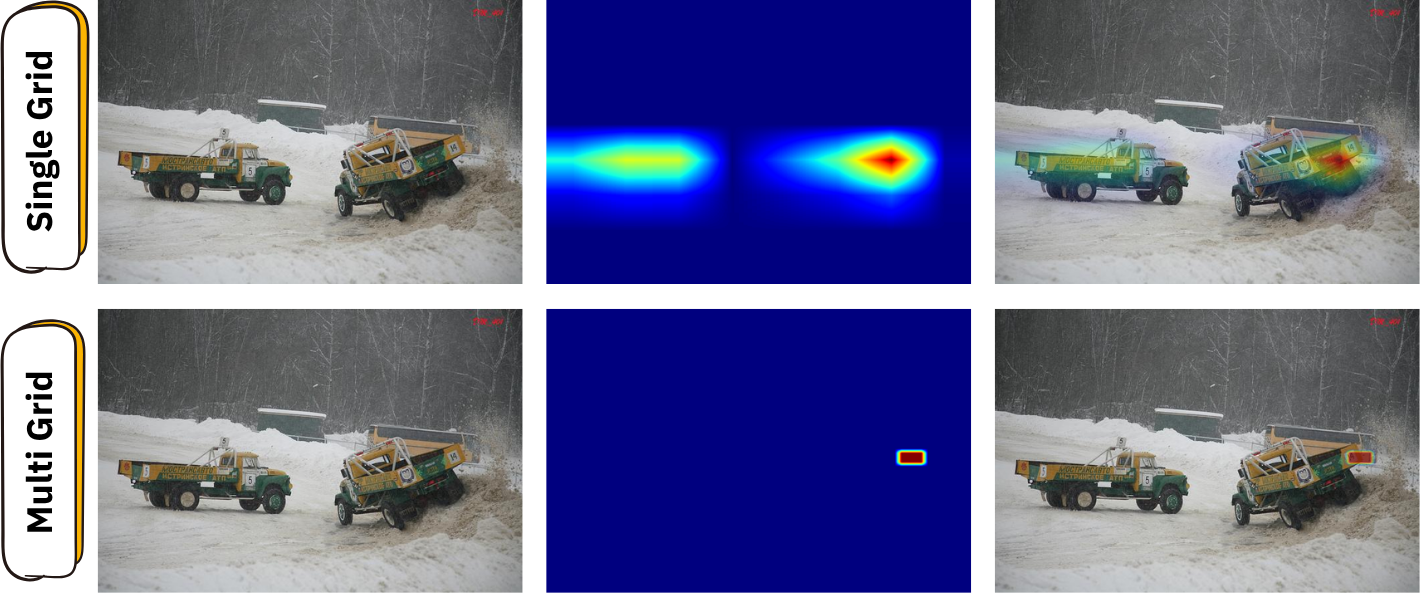}
\caption{\method{} on GPT-6-sol through its API, Query: "a truck number 14 on a snow bank." Logprobs only. Top: an
$8\times8$ grid, 16 calls. Bottom: the multigrid product,
58 calls, narrows the map onto the truck the query names.}
\label{fig:api_truck}
\end{figure}

Nothing in \method{} needs
more than first-token logprobs, so it runs unchanged on GPT-6-sol through
its API (Figures~\ref{fig:api_examples} and ~\ref{fig:api_truck}). An
$8\times8$ grid spreads the map over both trucks in the same rows, and the
multigrid product narrows it onto the one the query names. Two keyboards and two
yokes sharing a row are both lit, with nothing between them.

%% file: sections/applications.tex
\section{Read-outs as the Output Interface}\label{sec:applications}

The two tests established that the rationale is faithful. This section
shows that the read-out is general. The same map, with a different fixed
read-out on top, serves a different visual task, with no training and no
change to the operator, and each read-out is scored against ground truth
rather than against the model itself. 

\textbf{Read-out 1, the expectation: a location where pointing fails.}\label{sec:app:loc}
Generation is one interface to a model's spatial belief, and it can be
narrower than the belief. Table~\ref{tab:recovery} reads the belief
through the expectation and sets it against what the interface emits, on
a strong pointer, Qwen3-VL-4B, and on a weak one, the medical Lingshu-7B
\citep{lingshu2506}, whose pointing fails to parse on 30\% of RefCOCOg
queries. On the strong pointer generation is the richer channel and the
map recovers most of it. On Lingshu-7B the order flips and the map beats
the model's own point by 11 points. The belief is intact, the interface is
the bottleneck, and the gap between them is measurable.

\begin{table}[t]
\centering
\small
\caption{Does the map localize where the model's own pointing fails?
Point-in-region accuracy (\%) against ground-truth boxes, parse failures
scored as misses after getting retried 3 times. Qwen3-VL-4B on CAVE and RefCOCOg, and
the medical Lingshu-7B on RefCOCOg.}
\label{tab:recovery}
\begin{tabular}{lccc}
\toprule
 & \multicolumn{2}{c}{Qwen3-VL-4B} & Lingshu-7B \\
\textbf{Read-out} & CAVE & RefCOCOg & RefCOCOg \\
\midrule
Generated point (the interface) & \textbf{77.8} & \textbf{94.5} & 46.7 \\
Probe expectation (\method{})     & 71.3 & 67.5 & \textbf{57.7} \\
Attention (best layer)          & 57.2 & 40.0 & 40.0 \\
\midrule
Image-centre prior              & 58.1 & 40.5 & 39.3 \\
Random point                    & 24.3 & 24.8 & 24.3 \\
\bottomrule
\end{tabular}
\end{table}

\textbf{Read-out 2, the maximum: a hallucinated claim is anchored nowhere.}\label{sec:app:grounding}
The maximum reads how strongly the model commits to a region, and the
natural place to read it is a claim about object existence. POPE
\citep{pope2023} asks whether a named object is in the image, with labels
for which objects are present. A yes on a present object is a grounded
claim, a yes on an absent one is a hallucinated claim. On the full
benchmark the model says yes 4{,}006 times, 223 of them hallucinated, and
we read the map's maximum before it answers. A hallucinated object has
nowhere to peak, so the maximum separates the two kinds of claim at
ROC-AUC 0.833, with no labels and 16 queries. Deletion confirms the
reading from a second angle (Table~\ref{tab:halluc}). A grounded yes dies
only when its own region is deleted. A hallucinated yes breaks eight times
more often than a grounded one under a random deletion, and far more often
under deletion of its map region. 

\textbf{Read-out 3, the top-mass region: a crop the model can act on.}\label{sec:app:transfer}
A faithful explanation should be useful to the model itself, and a
confabulated one should not. We test this by acting on the map: probe
first, then crop the top-mass region from the full-resolution image and
append it as a close-up before answering. On TextVQA
(Table~\ref{tab:selfcond}) the map's crop lifts accuracy where a random
crop through the same mechanics lowers it. Behind the net numbers, the
map's crop turns 50\% of the model's 42 wrong answers right and the random
crop 38\% of them, while disturbing 4\% against 6\% of the answers the
model had right. 

\begin{table}[t]
\begin{minipage}[t]{0.48\linewidth}
\centering
\small
\caption{Does deletion tell a grounded yes from a hallucinated one? Flip
rate (\%) of yes claims on POPE when the map's region or a random region
is deleted.}
\label{tab:halluc}
\begin{tabular}{lcc}
\toprule
\textbf{Yes claim} & Map region & Random \\
\midrule
Grounded      & 41.0 & 1.4 \\
Hallucinated  & 69.5 & 11.2 \\
\bottomrule
\end{tabular}
\end{minipage}\hfill
\begin{minipage}[t]{0.48\linewidth}
\centering
\small
\caption{Does a crop of the map's region help the model? Accuracy (\%) on
TextVQA, n=500, Qwen3-VL-4B.}
\label{tab:selfcond}
\begin{tabular}{lc}
\toprule
\textbf{Added crop} & Accuracy \\
\midrule
None (image alone) & 91.6 \\
Own map region     & 92.4 \\
Random region      & 89.0 \\
\bottomrule
\end{tabular}
\end{minipage}
\end{table}

%% file: sections/ablations.tex
\section{Ablation Study}\label{sec:ablations}

All ablations use Qwen3-VL-4B on RefCOCOg. Table~\ref{tab:ablation} varies how a query budget
is spent. Both strategies trace an inverted U (Figure~\ref{fig:budget},
Appendix~\ref{app:figs}), because a band question is only informative
while the band retains enough context for an easy yes/no. Refining one
grid peaks narrowly at $K{=}6$ and collapses as bands thin. Composing
coprime grids reaches the same peak at $\{3,5\}$ and stays flat across a
five-fold budget range, so the value of composition is robustness. The
nested set $\{2,4,8\}$ spends more queries for less, because nested grids
double-count correlated evidence. The
gain is agreement-side: on deletion $\{3,5\}$ ties $K{=}8$ (52.3 against
53.4), since both select nearly the same top-mass cells.
Finally, the fusion rule behaves as the
product-of-experts reading predicts: product beats mean beats max. 

\begin{table}[t]
\centering
\small
\caption{How should a query budget be spent? NSS at the model's own point
on a fixed 296-example RefCOCOg split, Qwen3-VL-4B.}
\label{tab:ablation}
\setlength{\tabcolsep}{5pt}
\begin{tabular}{llcc}
\toprule
\textbf{Axis} & \textbf{Configuration} & Queries & NSS \\
\midrule
Refine one grid    & $K{=}2$                    & 4  & 0.75 \\
                   & $K{=}4$                    & 8  & 1.25 \\
                   & $K{=}6$                    & 12 & \textbf{1.46} \\
                   & $K{=}8$ (default)          & 16 & 1.14 \\
                   & $K{=}17$                   & 34 & 1.11 \\
                   & $K{=}40$                   & 80 & 0.22 \\
\midrule
Compose grids      & $\{3,5\}$, coprime         & 16 & \textbf{1.50} \\
                   & $\{2,3,5\}$, coprime       & 20 & \textbf{1.50} \\
                   & $\{2,4,8\}$, nested        & 28 & 1.25 \\
                   & $\{2,3,5,7,11\}$, coprime  & 56 & 1.25 \\
                   & $\{2,3,5,7,11,13\}$, coprime & 82 & 1.27 \\
\midrule
Fusion rule, $\{2,3,5\}$ & product              & 20 & \textbf{1.50} \\
                   & mean                       & 20 & 1.39 \\
                   & max                        & 20 & 0.89 \\
\bottomrule
\end{tabular}
\end{table}



%% file: sections/conclusion.tex
\section{Conclusion}\label{sec:conclusion}\label{sec:limitations}

We introduced \method{}, a training-free visual rationale read from the
output head of a frozen VLM: $2K$ yes/no band questions and an outer
product give a query-conditioned map, and a fixed read-out on the map
derives the answer instead of generating it, continuous and
image-dependent by construction. The map passes two ground-truth-free
faithfulness tests across three query distributions, four models, and two
families, where attention fails both and the strongest white-box baseline
trails, and three read-outs on the same map flag hallucinated objects,
localize where the model's own pointing fails, and fix half of the model's
wrong answers through a crop.

A few limitations and natural refinements remain. The operator needs a
model that accepts non-square inputs, since each band is a long thin
strip. And because the map is built from row and column answers, it knows
which rows and which columns contain the query but not which pairs of them
do: when the query matches two separate objects, the map also lights the
two empty cells where their rows and columns cross. Beyond these, the
read-out is the main design surface, and three steps follow: richer
read-outs on the same map (extent, count, relations), training on top of
the map as label-free supervision for the interface that cannot express
it, and probe-guided generation that acts on the audit before the claim is
generated. The answer posterior read through a fixed function is a new
axis for visual tasks, not yet optimized, and we expect the interface to
matter as much as the map.

%% file: sections/appendix.tex
\section{Appendix}

\subsection{Additional Figures and Tables}
\label{app:figs}

This appendix collects the figures the main text refers to and the cost
measurements behind Section~\ref{sec:ablations}.

\begin{figure}[h]
\centering
\includegraphics[width=0.9\linewidth]{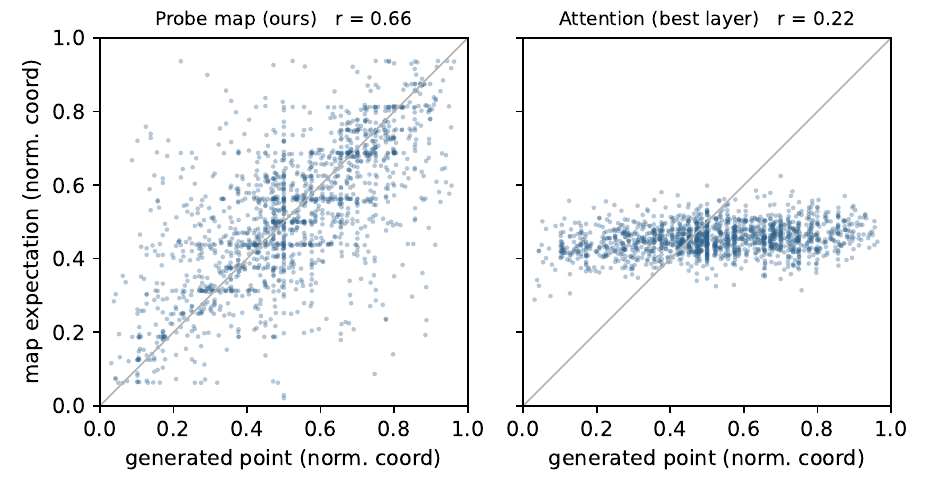}
\caption{Map expectation against the model's generated point on RefCOCOg.
The probe tracks the diagonal, best-layer attention is a flat band regardless
of where the model points.}
\label{fig:scatter}
\end{figure}

\paragraph{Why attention sits below chance.} Figure~\ref{fig:scatter}
plots each map's expectation against the model's generated point on
RefCOCOg. The probe's expectation tracks the point along the diagonal.
Best-layer attention's expectation is a flat band: its mass sits on sink
tokens irrelevant to the query, so it barely moves with the point, which
is what a below-chance AUC looks like in coordinates.

\begin{figure}[h]
\centering
\includegraphics[width=0.75\linewidth]{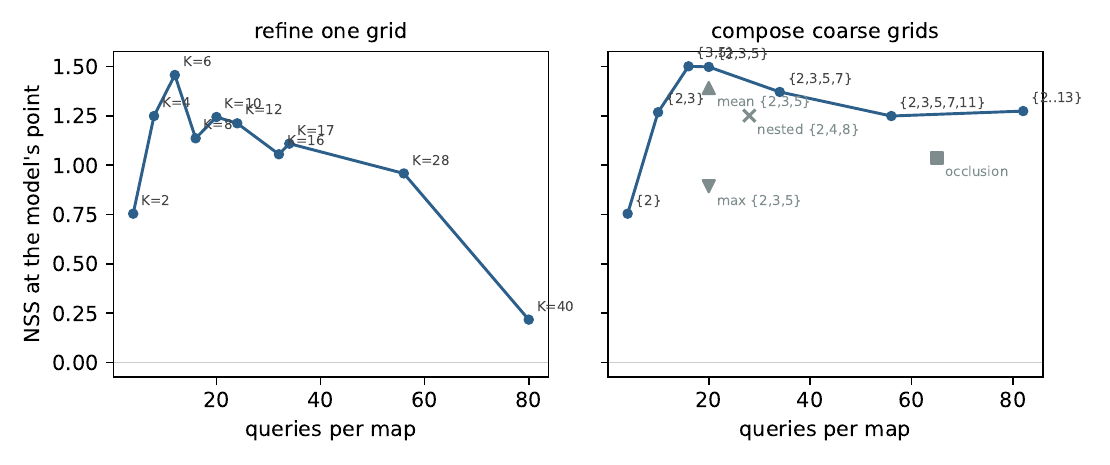}
\caption{Faithfulness against query budget. Refining one grid (left) peaks
narrowly and collapses, composing coprime grids (right) is flat-topped.
Grey markers, the fusion and independence controls of
Table~\ref{tab:ablation}.}
\label{fig:budget}
\end{figure}

\paragraph{Query budget curves.} Figure~\ref{fig:budget} plots
Table~\ref{tab:ablation} against the query budget. Refining one grid rises
to a narrow peak at $K{=}6$ and collapses as bands thin. Composing coprime
grids reaches the same peak and stays flat across a five-fold budget
range. The grey markers are the fusion and independence controls, mean,
max, and the nested set.

\paragraph{Cost.} The probe runs at 0.4 to 1.3 seconds per map from 4B to
30B, four to five times cheaper than occlusion, the other black-box method
(Table~\ref{app:cost}). Internal read-outs are cheaper per map where
the internals are available, since one forward pass beats 16, so the case
for the probe is access and faithfulness at modest cost, not raw speed.

\begin{table}[h]
\centering
\small
\caption{Seconds per explanation map on one A100, 20 samples at image side
512. Bold: the faster of the two black-box methods. T-MM additionally
carries gradient memory, up to 47\% extra at 4B.}
\label{app:cost}
\begin{tabular}{lcccc}
\toprule
\textbf{Method} & Queries & Access & 4B & 30B \\
\midrule
Probe map (ours) & 16 & logits only & \textbf{0.41} & \textbf{1.25} \\
Occlusion        & 65 & logits only & 1.80 & 5.90 \\
Attention (best layer) & 1 & internals, eager & 0.07 & 0.38 \\
Attention rollout & 1 & internals, eager & 0.59 & 1.15 \\
Relevancy T-MM   & 1 & internals, eager, gradients & 0.18 & 1.35 \\
\bottomrule
\end{tabular}
\end{table}

\subsection{Prompt Templates}
\label{app:prompts}

All prompts are listed verbatim. Placeholders in braces are filled at run
time, \texttt{\{cond\}} and \texttt{\{q\}} with the user query exactly as it
appears in the benchmark. In every prompt the images precede the text in
the model's chat template, and where two images are given, the full image
comes first and the crop second.

\paragraph{Band probe (every probe query).}
Each of the $2K$ band questions shows the model the band content and asks:
\begin{verbatim}
You are shown one or more adjacent tiles cropped from a larger
image. Is the following present in ANY of these tiles?
"{cond}"
Answer with exactly one word: Yes or No.
\end{verbatim}
The answer is never read from generated text. We take the first
answer-token logits and compute the yes posterior over the closed label
set, aggregating surface forms by logsumexp, yes forms
\texttt{\{Yes, yes, YES\}} with and without a leading space, and likewise
for no.

\paragraph{Pointing reference (agreement test and point-in-region).}
The generated point uses each family's native grounding convention,
declared per family rather than inferred, coordinates are parsed from the
reply and interpreted on a 0 to 1000 scale for both families below.
Qwen3-VL models, and medical derivatives built on Qwen-VL such as Lingshu:
\begin{verbatim}
Locate "{q}" in this image and output its center point as
JSON: {"point_2d": [x, y], "label": "{q}"}
\end{verbatim}
InternVL models, the documented grounding template, the box centre is
taken as the point:
\begin{verbatim}
Please provide the bounding box coordinate of the region this
sentence describes: <ref>{q}</ref>
\end{verbatim}
Other models, a generic pixel-space prompt:
\begin{verbatim}
Point to {q} in this image. Output ONLY the pixel coordinates
of its center as (x, y). No other text.
\end{verbatim}

\paragraph{Open-ended answering (deletion test, image-reliance labeling,
self-conditioning).}
\begin{verbatim}
{q}
Answer with a single word or phrase.
\end{verbatim}

\paragraph{Existence claims (hallucination audit).}
\begin{verbatim}
{q}
Answer yes or no.
\end{verbatim}

\paragraph{Self-conditioning close-up.}
The receiver sees the full image and the crop of the map's top-mass region
as a second image:
\begin{verbatim}
{q}
The second image is a close-up of the region most relevant to
the question.
Answer with a single word or phrase.
\end{verbatim}

\subsection{Qualitative Examples}
\label{app:qualitative}

\begin{figure}[h]
    \centering
    \includegraphics[width=0.7\linewidth]{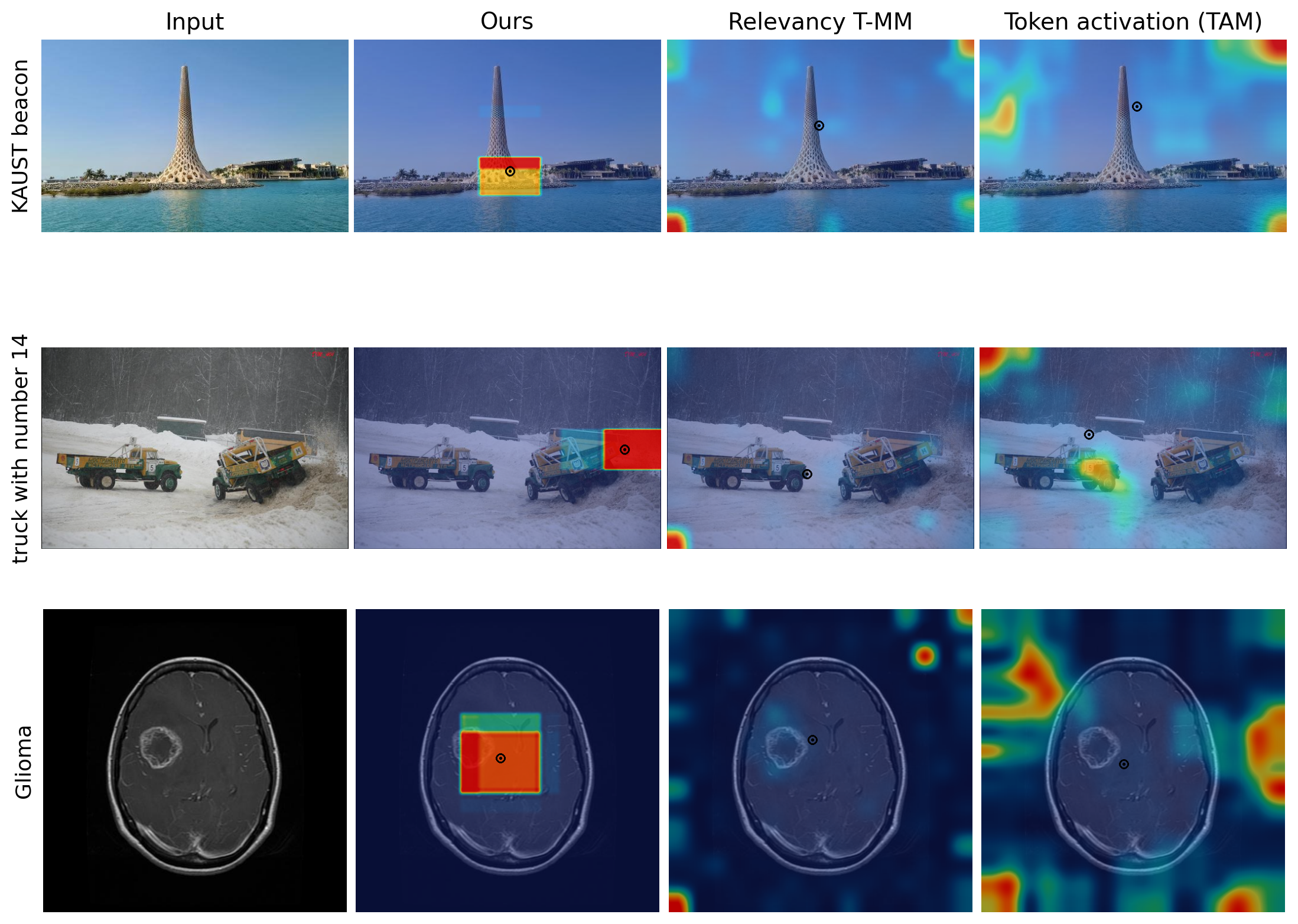}
    \caption{Qualitative examples of the probe map. Each column shows one
example, with the input image above its corresponding probe map.}
\label{fig:qualitative}
\end{figure}

Figure~\ref{fig:qualitative} shows three probe maps at $K{=}8$ on
Qwen3-VL-4B, each with its query as the sub-caption: a landmark, a
referring expression with a distractor, and a pathology slide with a
medical query. In each the map is a single mode on the queried region,
and the third shows that the operator works even on specific domains, the
band question carries the query as is.

\subsection{Additional Diagnostics}
\label{app:diag}

This appendix holds the diagnostics behind Section~\ref{sec:results}: the
attention layer sweep, the image-reliance split of the deletion test, the
full four-metric agreement table with the control rows, the map-maximum
grading, and the localization-heads variant.

\begin{figure}[h]
\centering
\includegraphics[width=0.7\linewidth]{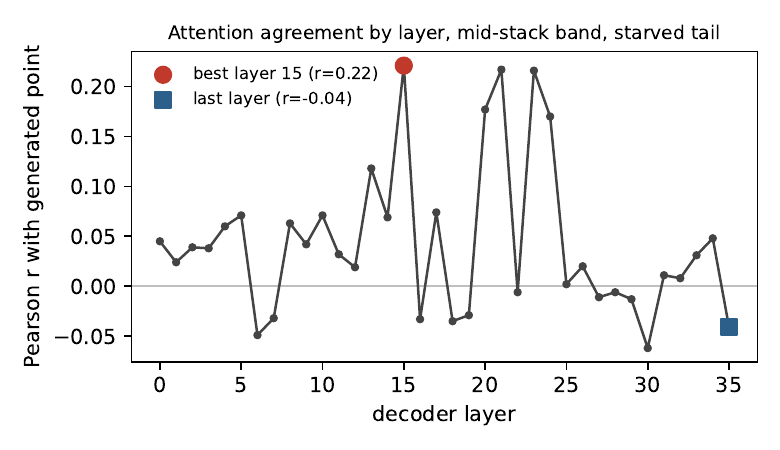}
\caption{Agreement of each decoder layer's attention map with the model's
point on Qwen3-VL-4B. A mid-stack band peaks at layer 15, the last layer is
negative.}
\label{fig:layers}
\end{figure}

\paragraph{Attention layer sweep.} Figure~\ref{fig:layers} scores the
attention map of every decoder layer of Qwen3-VL-4B against the model's
generated point. A mid-stack band peaks at layer 15, the layer every
attention row in the main tables uses, and the last layer is negative.

\paragraph{Where the map matters.} Whole-image deletion labels each
correctly answered question by whether the image was needed. Probe-region
deletion flips 57.4\% of the questions that needed the image and 22.0\% of
those that did not, and the split holds at every scale and on both
families (Table~\ref{tab:reliance}).

\begin{table}[h]
\centering
\small
\caption{The map matters where the image mattered. Probe-region flip rate
(\%) on questions the image was needed for against questions it was not,
labeled by whole-image deletion. TextVQA unless noted.}
\label{tab:reliance}
\begin{tabular}{lccc}
\toprule
\textbf{Model} & Image used & Image unused & Random (unused) \\
\midrule
Qwen3-VL-4B            & \textbf{57.4} & 22.0 & 2.4 \\
Qwen3-VL-8B            & \textbf{53.8} & 31.1 & 2.2 \\
Qwen3-VL-30B-A3B       & \textbf{48.4} & 22.2 & 0.0 \\
InternVL3.5-8B         & \textbf{54.5} & 20.6 & 5.9 \\
GQA non-binary, 4B     & \textbf{30.0} & 18.7 & 0.0 \\
\bottomrule
\end{tabular}
\end{table}

\paragraph{Agreement in all four metrics.} Table~\ref{tab:agreement_detail}
adds to Table~\ref{tab:agreement} the distance from the map's expectation
to the generated point, as a fraction of the image diagonal, and the
Pearson correlation of their coordinates, on RefCOCOg. The image-centre
prior is the floor for the distance. The two control rows show the
guarantee of Section~\ref{sec:method:op} in coordinates: on a blank image
the expectation sits exactly on the centre, and on a swapped image it
lands farther from the point than the centre does, with no correlation, so
the map follows the image it was shown.

\begin{table}[h]
\centering
\caption{Full agreement detail on RefCOCOg, all four metrics. Bold: best
per column among the methods.}
\label{tab:agreement_detail}
\small
\begin{tabular}{lcccc}
\toprule
Method & NSS & AUC & Dist.\ $\downarrow$ & Pearson $r$ \\
\midrule
Probe map ($K{=}8$)        & 1.36 & 0.824 & 0.130 & 0.665 \\
Probe map ($\{3,5\}$)      & \textbf{1.56} & \textbf{0.853} & \textbf{0.122} & \textbf{0.711} \\
Attention (best layer, $\ell{=}15$) & $-0.22$ & 0.380 & 0.193 & 0.224 \\
Attention (last layer)     & $-0.15$ & 0.464 & 0.193 & $-0.036$ \\
Attention rollout          & $-0.16$ & 0.457 & 0.292 & 0.120 \\
Relevancy T-MM (all layers) & 0.33 & 0.656 & 0.159 & 0.706 \\
Relevancy T-MM (last block) & $-0.15$ & 0.420 & 0.213 & 0.048 \\
Occlusion                  & 1.04 & 0.649 & 0.160 & 0.504 \\
Token activation TAM (own regime) & 0.77 & 0.693 & 0.179 & 0.446 \\
Random map                 & $-0.05$ & 0.486 & 0.188 & 0.048 \\
Probe map, blank image     & 0.00    & 0.500 & 0.189 & -- \\
Probe map, swapped image   & $-0.02$ & 0.475 & 0.224 & $-0.055$ \\
Image-centre prior         & --      & --    & 0.189 & -- \\
\bottomrule
\end{tabular}
\end{table}

\paragraph{The maximum grades causal strength.} Binning by the map
maximum, the probe-region flip rate rises from the lowest to the highest
tercile in all five configurations tested (Table~\ref{tab:grading}), and
as a scalar the maximum predicts the flip with AUC up to 0.677 on the 30B.
The maximum reads spatial commitment, not answer reliance: it does not
predict whether the answer needed the image at all (AUC 0.49 against the
whole-image label), which is the label's question, not the map's.

\begin{table}[h]
\centering
\caption{Map-maximum grading. Probe-region flip rate (\%) on the lowest
and highest tercile of the map maximum, per configuration.}
\label{tab:grading}
\small
\begin{tabular}{lcc}
\toprule
Configuration & Low tercile & High tercile \\
\midrule
TextVQA, 4B, blank          & 44.3 & \textbf{64.2} \\
TextVQA, 4B, blur           & 49.2 & \textbf{61.8} \\
GQA, 4B                     & 16.1 & \textbf{28.4} \\
TextVQA, 30B                & 26.9 & \textbf{58.5} \\
TextVQA, InternVL3.5-8B     & 48.3 & \textbf{60.7} \\
\bottomrule
\end{tabular}
\end{table}

\paragraph{Localization heads, per-example variant.}
The published method selects heads on a calibration corpus. A per-example
adaptation with no calibration performs at noise level on both tests
(agreement distance 0.375, coordinate correlation $-0.051$ on RefCOCOg,
deletion at the random control), consistent with the authors' own finding
that uncalibrated attention maps are sparse and noisy
\citep{attnheads2503}.

\subsection{Configuration Details}
\label{app:config}

Table~\ref{tab:config} lists every fixed setting, and the protocol notes
below cover scoring and the cases the main text does not spell out.

\paragraph{Protocol notes.} All runs use one NVIDIA A100 80GB with
HuggingFace Transformers. Open-ended answers are scored by normalized
string match against the annotator answers. In point-in-region
evaluation a parse failure is scored as a miss. Random regions and random
maps use fixed per-example seeds. In Test 1 each map is scored at its
native resolution, and the finest map, the multigrid product, scores
highest, so resolution does not drive the ordering. The probe's own
resolution does not either: on a fixed 300-example RefCOCOg split the
$K{=}8$ probe scores NSS 1.14 and AUC 0.801 at a 512-pixel long side and
1.14 and 0.807 at 1024, the resolution the reference point is generated
at.

\begin{table}[h]
\centering
\caption{Per-experiment configuration. All values are fixed across models
and benchmarks unless a sweep is the experiment.}
\label{tab:config}
\small
\begin{tabular}{lp{0.6\linewidth}}
\toprule
Probe image long side & 512 px \\
Answering and deletion long side & 1024 px \\
Self-conditioning long side & 1280 px (the crop inherits it) \\
Grid & $K{=}8$ default, multigrid product $\{3,5\}$ or $\{2,3,5\}$ \\
Multigrid shared grid & $64\times64$, nearest-neighbour upsampling \\
Band presentation & one concatenated strip image per question \\
Deletion region & top-mass cells, 12\% of the grid (8 of 64) \\
Deletion fill & grey RGB (127,127,127), blur variant Gaussian \\
Random control & same cell count, disjoint from every method region \\
Crop (self-conditioning) & bounding box of the 8 top-mass cells, expanded 18\% per side, minimum 96 px \\
Attention layer selection & swept over all decoder layers on the first 200 RefCOCOg examples, once per model, then fixed \\
Decoding & greedy, deterministic, fixed per-example seeds \\
\bottomrule
\end{tabular}
\end{table}